\documentclass[journal,twocolumn]{IEEEtran}
\usepackage{amsmath,amsfonts}
\usepackage{algorithmic}
\usepackage{algorithm}
\usepackage{array}
\usepackage{textcomp}
\usepackage{stfloats}
\usepackage{url}
\usepackage{verbatim}
\usepackage{graphicx}
\usepackage{epstopdf}
\usepackage{cite}
\usepackage{booktabs}
\usepackage{threeparttable}
\usepackage{multirow}
\usepackage{xspace}
\usepackage{newtxtext,newtxmath}

\def\BibTeX{{\rm B\kern-.05em{\sc i\kern-.025em b}\kern-.08em
        T\kern-.1667em\lower.7ex\hbox{E}\kern-.125emX}}
\newcommand{\R}{\mathbb{R}}
\newcommand{\model}{ISCL\xspace}
\newtheorem{definition}{Definition}
\newtheorem{theorem}{Theorem}

\newtheorem{lemma}{Lemma}

\makeatletter
\def\@begintheorem#1#2{\@IEEEtmpitemindent\itemindent\topsep 0pt\rmfamily\trivlist%
    \item[\hskip \labelsep{\indent\bfseries #1\ #2:}]\itemindent\@IEEEtmpitemindent}
\def\@opargbegintheorem#1#2#3{\@IEEEtmpitemindent\itemindent\topsep 0pt\rmfamily \trivlist%
    \item[\hskip\labelsep{\indent\bfseries #1\ #2\ (#3):}]\itemindent\@IEEEtmpitemindent}
\makeatother
\graphicspath{{figures/}}

\begin{document}

\title{Identifiable Signed Causal Learning via Latent Stochastic Differential Equations for Inferring Brain Dynamic Effective Connectivity}

\author{Yiding Wang, Chen Qiao
    \thanks{This work has been submitted to the IEEE for possible publication. Copyright may be transferred without notice, after which this version may no longer be accessible.}
    \thanks{This research was supported by the National Natural Science Foundation of China (Nos. 12271429 and 12226007), and the National Key R\&D Program of China (No. 2024YFA1012000). (Corresponding author: Chen Qiao)}
    \thanks{Yiding Wang and Chen Qiao are with the School of Mathematics and Statistics, Xi'an Jiaotong University, Xi'an 710049 P.R. China. (ytcur1@stu.xjtu.edu.cn; qiaochen@mail.xjtu.edu.cn)}}

\maketitle

\begin{abstract}
Understanding how brain regions interact to support cognitive tasks requires characterizing directed causal influences among them, known as effective connectivity (EC).
As inter-regional coupling changes over time, a dynamic EC (dEC) estimate is required to track this temporal variation.
Existing dEC learning methods typically lack a causal rationality guarantee or an identifiability analysis, and output unsigned connectivity that cannot distinguish excitatory from inhibitory influences, which limits further study of brain mechanisms.
We propose \model, a latent stochastic differential equation framework that derives dEC from latent dynamics through a causal influence operator defined in observation space.
As established by a rationality theorem, the operator captures a directed interventional response at each brain state, and under invariance assumptions the operator is identifiable from data, with its signed output separating excitatory from inhibitory influences.
On synthetic benchmarks, \model recovers the causal structure on Lorenz96 and achieves the lowest mean structural Hamming distance among compared methods on NetSim, while preserving signed influence structures that distinguish excitatory from inhibitory interactions.
Applied to HCP Social Cognition task-fMRI, \model reveals a signed architecture whose stability across seeds supports the identifiability guarantee, identifies dominant source and target networks consistent with social-cognition systems, and shows that condition-specific training strengthens causal outflow from the default-mode and frontoparietal networks during theory-of-mind processing.
In this task, the recovered signed structure reveals network-level excitatory and inhibitory organization, with the dorsal-attention network as the only consistently positive out-strength source.
The source code of the proposed model is available at https://github.com/Yiding00/ISCL.
\end{abstract}


\section{Introduction}
\IEEEPARstart{B}{rain} function depends on interactions among cortical regions, and recovering these interactions is a core problem in network neuroscience~\cite{bullmore2009complex}.
Neuroimaging methods such as functional magnetic resonance imaging (fMRI) enable non-invasive observation of large-scale brain activity~\cite{biswal2010toward}.
A central goal in network neuroscience is to estimate directed influences among brain regions, commonly referred to as effective connectivity (EC)~\cite{friston2013analysing}.
Brain dynamics are inherently time-varying.
Cognitive demands, attentional shifts, and transient neural events reshape inter-regional coupling on sub-second timescales~\cite{calhoun2014chronnectome}.
Dynamic effective connectivity (dEC) that captures this temporal variation is needed to track how directed influences evolve over a task or resting-state scan.

Classical EC methods such as dynamic causal modeling (DCM)~\cite{friston2003dynamic} and approaches based on Granger causality (GC)~\cite{shojaie2022granger} are interpretable causal learning methods, but do not scale well to high-dimensional data~\cite{friston2013analysing}.
DCM requires candidate models to be specified in advance and compares them through Bayesian model selection~\cite{friston2003dynamic}.
Practically, classical DCM is limited to small networks by the combinatorial cost of exploring the model space and by Bayesian inversion~\cite{stephan2010ten}, and EC estimates show only moderate test-retest reliability~\cite{almgren2018variability}.
GC methods use temporal precedence and improved prediction as evidence of directionality, but their stationarity and linearity assumptions are hard to maintain for high-dimensional brain signals with nonlinear dynamics~\cite{shojaie2022granger}.

Deep architectures have been introduced to relax these constraints~\cite{tank2021neural,xu2025deep,liu2024mcan,li2022virtual,wang2024deep,lu2025netformer,wang2026time,wang2026idecdct}.
Neural Granger causality methods extend the Granger framework with recurrent or convolutional architectures to model nonlinear dependencies~\cite{tank2021neural,xu2025deep}.
Generative adversarial frameworks recover effective connectivity from multimodal brain data~\cite{liu2024mcan}.
Deep feature-aggregation networks built on dynamic EC estimates identify clinical conditions~\cite{li2022virtual}, and deep causal learners based on structural equation models estimate time-varying causal masks under sparsity regularization~\cite{wang2024deep}.
A recent model inspired by transformers recovers time-varying inter-neuron connectivity by treating the linearized attention matrix itself as the coupling weight~\cite{lu2025netformer}.
Time-reversal methods exploit temporal asymmetries to infer directed graphs~\cite{wang2026time}.
A continuous-time framework learns identifiable dynamic EC through causal distribution modeling~\cite{wang2026idecdct}.
These approaches have advanced directed connectivity estimation and scale to high-dimensional signals that classical methods cannot reach.

Continuous-time models of neural activity fall into two branches.
One branch performs continuous-time structure learning at the variable level, without a low-dimensional latent projection~\cite{bellot2022neural,wang2024neural}.
Score-based methods estimate directed graphs by fitting penalized neural ordinary differential equations with adaptive regularization~\cite{bellot2022neural}, and variational approaches place a posterior distribution over possible graphs while the dynamics are modeled by neural stochastic differential equations (SDEs)~\cite{wang2024neural}.
Both recover a discrete graph of which variables interact.
A second branch uses a low-dimensional latent SDE with a decoder that maps latent states to high-dimensional observations, and establishes that the latent SDE is identifiable up to an isometry~\cite{hasan2021identifying}.
The first operates without a latent space, and the second leaves connectivity in latent coordinates without a direct mapping to region-level causal influence.
Neither branch recovers continuous coupling at the region level in observation space using latent dynamics.

A reliable dEC estimate must satisfy three properties that existing methods do not jointly provide.
Rationality ensures that the estimated dEC measures a directed interventional response at a given brain state.
Identifiability ensures that two different objects cannot induce the same observation distribution, so that the learned dEC is uniquely determined by the data and not by an arbitrary parameterization.
A signed output preserves the sign of coupling, distinguishing excitatory from inhibitory influences.
Together, these properties support consistency across runs and provide a reliable basis for cross-subject comparison and mechanistic interpretation of brain function.

To achieve these properties, we propose Identifiable Signed Causal Learning (\model), a latent SDE framework that derives dEC from learned latent dynamics through a causal influence operator defined in observation space.
The latent-to-observation structure links the model to the blood oxygen level dependent (BOLD) signal generation process, mapping high-dimensional observations into a low-dimensional latent space that models the underlying neural dynamics.
The main contributions are summarized as follows.
\begin{itemize}
    \item A continuous-time dynamic causal learning framework is introduced that models neural dynamics through a latent SDE and a decoder mapping latent states to observations, producing a causal influence matrix at each time point from high-dimensional observations.
    The computationally fast observation-space causal influence operator is analytically derived from the trained decoder and drift, and a rationality theorem establishes that it captures a directed interventional response at each brain state.
    \item Under invariance assumptions of the latent SDE, the operator is identifiable in observation space, so the reported connectivity does not depend on the particular latent parameterization.
    On synthetic benchmarks and task-fMRI, the recovered connectivity is consistent across seeds, supporting the identifiability guarantee.
    \item The operator outputs signed region-level dEC, preserving the distinction between excitatory and inhibitory influences.
    On HCP, the signed structure reveals the dorsal-attention network as the only consistently positive out-strength source and characterizes EC reconfiguration across task states, with causal outflow from the default-mode and frontoparietal networks strengthened during theory-of-mind processing.
\end{itemize}

The remainder of this paper is organized as follows.
Section~\ref{sec:methodology} presents the \model framework and the identifiable causal influence operator.
Section~\ref{sec:experiments} reports experiments on synthetic benchmarks and HCP Social Cognition task-fMRI.
Section~\ref{sec:conclusion} concludes the paper.

\section{Methodology}
\label{sec:methodology}

\begin{figure*}[t]
\centering
\includegraphics[width=\linewidth]{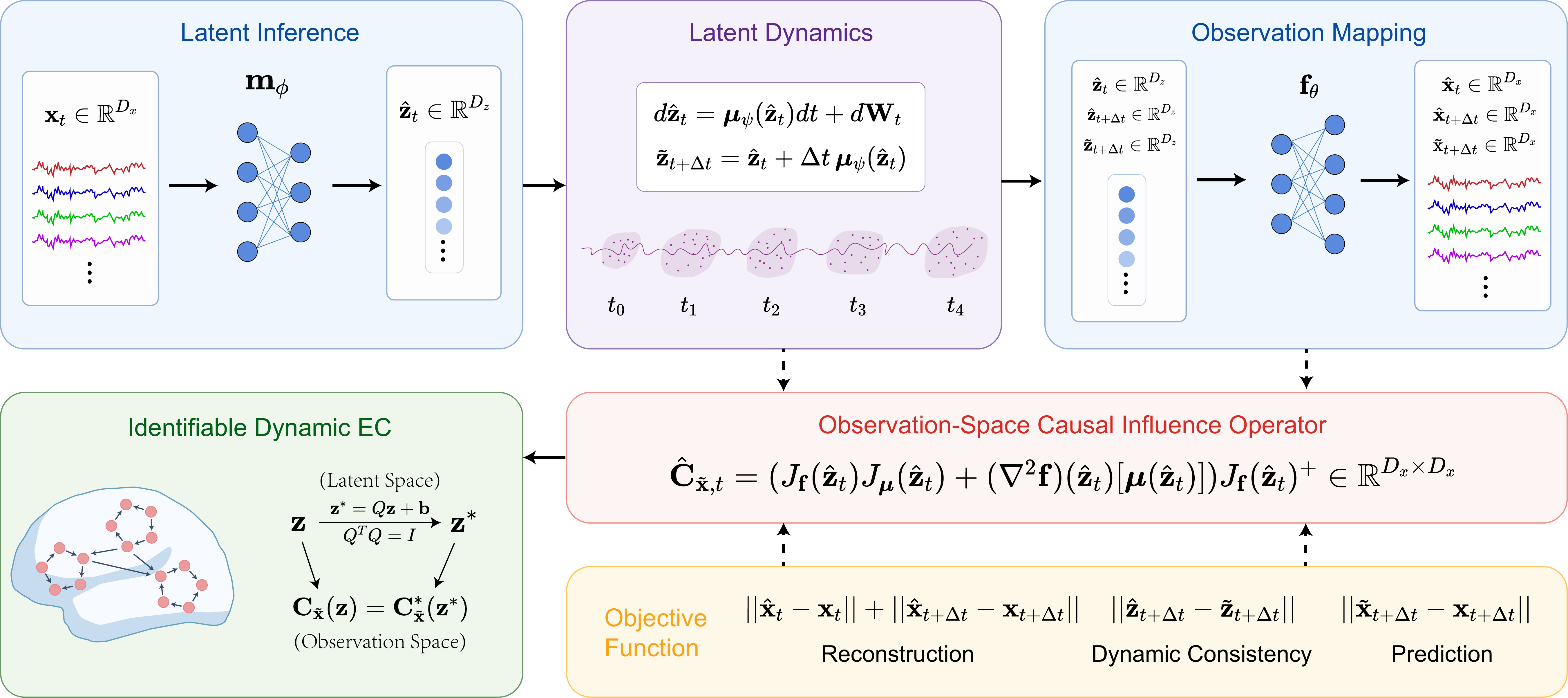}
\caption{Overview of the \model framework. High-dimensional observations $\mathbf{x}_t$ are mapped into a low-dimensional latent representation $\hat{\mathbf{z}}_t$ via latent inference with encoder $\mathbf{m}_{\phi}$. Temporal evolution is modeled by latent dynamics with drift $\boldsymbol{\mu}_\psi$, producing dynamically consistent latent trajectories. Observation mapping via decoder $\mathbf{f}_\theta$ reconstructs observations and predicts future states from the latent dynamics. The Jacobian $\nabla \mathbf{f}_\theta$ propagates latent drift to the observation space via the chain rule, and the Hessian $\nabla^2 \mathbf{f}_\theta$ provides a second-order correction, yielding an identifiable dynamic EC operator $\mathbf{C}_{\tilde{\mathbf{x}}} \in \R^{D_x \times D_x}$. Training jointly optimizes reconstruction accuracy, latent dynamic consistency, and future-state prediction.}
\label{fig:overview}
\end{figure*}

In this section, we present a latent-to-observation dynamic causal learning framework for revealing dEC with latent SDEs.
The overall framework is illustrated in Fig.~\ref{fig:overview}.
We define a latent dynamics model and the estimation problem, construct an observation-space causal influence operator, analyze its rationality and identifiability, describe the model architecture, and present the training objective.
\subsection{Latent Dynamics Modeling}

Let $\mathbf{x}_t \in \R^{D_x}$ be the observed signal at time $t$, where $D_x$ is the number of observational nodes, and let $\mathbf{z}_t \in \R^{D_z}$ be the corresponding latent state, where $D_z\le D_x$ is the latent dimension.
The goal is to recover a $D_x \times D_x$ directed causal influence matrix $\mathbf{C}_t$ in the observation space for each time point $t$.
Its $(i,j)$ entry encodes the observation-level causal influence $j \to i$, with column $j$ as the source and row $i$ as the target.
From a neurobiological perspective, a positive entry indicates an excitatory influence, where a tangent perturbation at node $j$ increases the BOLD velocity at node $i$.
A negative entry indicates an inhibitory influence, where the perturbation decreases the BOLD velocity.
Because $\mathbf{C}_t$ captures this signed, directed interventional response at each time point, it constitutes a dEC estimate.
This estimate is required to be invariant to the latent-coordinate ambiguity of the data-generating process, and to reflect the local response of node activity to infinitesimal perturbations at other nodes, not pairwise statistical association.
Thus, we introduce a latent-to-observation model.
The observation process is defined as
\begin{equation*}
\mathbf{x}_t = \mathbf{f}_{\theta}(\mathbf{z}_t) + \boldsymbol{\epsilon}_t
\end{equation*}
where $\mathbf{f}_{\theta}: \R^{D_z}\to\R^{D_x}$ is a smooth observation function parameterized by $\theta$, and $\boldsymbol{\epsilon}_t$ is zero-mean Gaussian noise.

Motivated by Hasan et al.~\cite{hasan2021identifying}, we consider a latent SDE with isotropic diffusion as continuous-time latent dynamics
\begin{equation*}
d\mathbf{z}_t = \boldsymbol{\mu}_{\psi}(\mathbf{z}_t)\,dt + d\mathbf{W}_t
\end{equation*}
where $\boldsymbol{\mu}_{\psi}: \R^{D_z} \to \R^{D_z}$ is the drift function parameterized by $\psi$, and $\mathbf{W}_t$ is a Wiener process.
The isotropic diffusion matches the identifiability conditions of Hasan et al.~\cite{hasan2021identifying}, under which both the latent SDE coefficients and the latent state are identifiable up to an isometry.
The learned drift $\boldsymbol{\mu}_\psi$ therefore captures interaction structure uniquely determined by the data.

Intuitively, the model follows the basic logic of fMRI signal generation.
The latent state $\mathbf{z}_t$ represents underlying neural activity at the network level, which evolves over time driven by the drift $\boldsymbol{\mu}_\psi$ and random fluctuations.
The decoder $\mathbf{f}_{\theta}$ acts as the hemodynamic forward model, mapping this hidden neural activity to the BOLD signals observed at each brain region.

\subsection{Observation-Space Causal Influence Operator}

In this section, we define the observation-space causal influence operator that maps latent dynamics to causal interactions.
The operator is defined on the deterministic drift skeleton of the latent SDE.
When two neighboring latent trajectories share the same Brownian path, subtracting their SDEs cancels the diffusion terms, and the first-order perturbation propagation depends only on the drift and the observation map.
Let
\begin{equation*}
\mathcal{M}=\mathrm{Im} (\mathbf{f}_{\theta}) \subset \R^{D_x}
\end{equation*}
denote the observation manifold, and let
\begin{equation*}
\tilde{\mathbf{x}} = \mathbf{f}_{\theta}(\mathbf{z})
\end{equation*}
denote the noise-free decoder output on $\mathcal{M}$, distinct from the noisy observation $\mathbf{x}_t$.
Tracing how the drift propagates through $\mathbf{f}_\theta$ leads to the following operator.

\begin{definition}
\label{def:obs_causal_operator}
Assume that $\mathbf{f}: \R^{D_z}\to \R^{D_x}$ is a smooth immersion and that $J_{\mathbf{f}}(\mathbf{z})$ has full column rank. The observation-space causal influence operator at $\tilde{\mathbf{x}}=\mathbf{f}(\mathbf{z})$ is defined as
\begin{equation}
\mathbf{C}_{\tilde{\mathbf{x}}}(\mathbf{z}) :=
\Big(
J_{\mathbf{f}}(\mathbf{z})\,J_{\boldsymbol{\mu}}(\mathbf{z})
+
(\nabla^2 \mathbf{f})(\mathbf{z})[\boldsymbol{\mu}(\mathbf{z})]
\Big)
J_{\mathbf{f}}(\mathbf{z})^{+}
\label{eq:obs_causal_operator}
\end{equation}
where $J_{\mathbf{f}}(\mathbf{z})^{+}$ denotes the Moore--Penrose pseudoinverse of the decoder Jacobian $J_{\mathbf{f}}(\mathbf{z})$~\cite{penrose1955generalized}, and $(\nabla^2 \mathbf{f})(\mathbf{z})[\boldsymbol{\mu}(\mathbf{z})]$ denotes the contraction of the decoder Hessian with the drift vector, defined componentwise as
\begin{equation*}
\big[(\nabla^2 \mathbf{f})(\mathbf{z})\big[\boldsymbol{\mu}(\mathbf{z})\big]\big]_{ij}=\sum_{k=1}^{D_z}
\frac{\partial^2 f_i(\mathbf{z})}{\partial z_j\partial z_k}\,\mu_k(\mathbf{z})
\end{equation*}
\end{definition}

Throughout this paper, excitatory and inhibitory refer to BOLD-velocity response directions, following the convention in DCM~\cite{friston2003dynamic} and other effective connectivity methods.
$\mathbf{C}_{\tilde{\mathbf{x}}}$ captures the first-order response of the observation velocity to tangent perturbations, and is reported as a $D_x \times D_x$ matrix in the original observation coordinates.

The following theorem formalizes this property.
\begin{theorem}[Rationality]
\label{thm:rationality}
Let $\dot{\tilde{\mathbf{x}}} = J_{\mathbf{f}}(\mathbf{z})\boldsymbol{\mu}(\mathbf{z})$ denote the velocity of $\tilde{\mathbf{x}}$.
Under the assumptions of Definition~\ref{def:obs_causal_operator}, for any tangent perturbation $d\mathbf{x}\in T_{\tilde{\mathbf{x}}}\mathcal{M}$, the induced local response of the observation velocity satisfies
\begin{equation*}
d\dot{\tilde{\mathbf{x}}} = \mathbf{C}_{\tilde{\mathbf{x}}}(\mathbf{z})\,d\mathbf{x}
\end{equation*}
\end{theorem}
\noindent\textit{Proof sketch.} The result follows from a chain-rule expansion of $\dot{\tilde{\mathbf{x}}}=J_{\mathbf{f}}(\mathbf{z})\boldsymbol{\mu}(\mathbf{z})$ and the full-column-rank substitution $d\mathbf{z}=J_{\mathbf{f}}(\mathbf{z})^{+}d\mathbf{x}$; the full proof is given in Appendix~\ref{app:identifiability-proofs}.
\hfill$\blacksquare$

When $D_z=D_x$ and the decoder is invertible, the tangent space $T_{\tilde{\mathbf{x}}}\mathcal{M}$ coincides with $\R^{D_x}$, and each column of $\mathbf{C}_{\tilde{\mathbf{x}}}$ directly represents the response to an independent perturbation at the corresponding observed node.
When $D_z<D_x$, the tangent space is a proper subspace, and $\mathbf{C}_{\tilde{\mathbf{x}}}$ describes the response only to perturbations lying in this subspace.
Column $j$ then represents the response to the tangent projection of a perturbation at node $j$, rather than to an independent coordinate intervention.

For an affine decoder $\mathbf{f}(\mathbf{z})=\mathbf{A}\mathbf{z}+\mathbf{b}$, the Hessian term vanishes and Eq.~(\ref{eq:obs_causal_operator}) reduces to
\begin{equation*}
\mathbf{C}_{\tilde{\mathbf{x}}}(\mathbf{z}) =
J_{\mathbf{f}}(\mathbf{z})\,J_{\boldsymbol{\mu}}(\mathbf{z})\,J_{\mathbf{f}}(\mathbf{z})^{+}
\end{equation*}
When the decoder is the identity map, as in the vector autoregressive model with $D_z=D_x$, the Hessian vanishes and $J_{\mathbf{f}}=J_{\mathbf{f}}^{+}=I$, so the operator reduces to $J_{\boldsymbol{\mu}}$, recovering the autoregressive coefficient matrix.

\subsection{Identifiability Analysis}
The following theorem states the identifiability of the observation-space causal operator under the common latent-SDE identifiability assumptions used in our analysis.

\begin{theorem}[Identifiability]
\label{thm:identifiability}
Assume that: (A1) the drift $\boldsymbol{\mu}$ and diffusion $\boldsymbol{\sigma}$ of the latent SDE are globally Lipschitz, $\boldsymbol{\sigma}$ is symmetric positive definite and reducible, meaning the SDE admits an invertible transform that reduces its diffusion to isotropic form, and the observation noise $\boldsymbol{\epsilon}$ is Gaussian with non-degenerate covariance; (A2) the observation map $\mathbf{f}$ is $C^2$ and $J_{\mathbf{f}}(\mathbf{z})$ has full column rank; and (A3) two parameterizations share the same latent dimension and isotropic diffusion. Let $(\mathbf{f},\boldsymbol{\mu})$ and $(\mathbf{f}^{\ast},\boldsymbol{\mu}^{\ast})$ be two parameterizations that induce the same joint distribution of adjacent observations $(\mathbf{x}_t,\mathbf{x}_{t+\Delta t})$. Then, for every common observation point $\tilde{\mathbf{x}}=\mathbf{f}(\mathbf{z})=\mathbf{f}^{\ast}(\mathbf{z}^{\ast})$,
\begin{equation*}
\mathbf{C}^{\ast}_{\tilde{\mathbf{x}}}(\mathbf{z}^{\ast})
=
\mathbf{C}_{\tilde{\mathbf{x}}}(\mathbf{z})
\end{equation*}
Thus, the observation-space causal operator is identifiable in observation coordinates up to isometric latent transformations.
\end{theorem}
\noindent\textit{Proof sketch.} Under assumptions (A1)--(A3), two models generating the same joint distribution differ at most by an isometric transformation $\mathbf{z}^{\ast} = Q\mathbf{z} + \mathbf{b}$ with $Q$ orthogonal.
Under this transformation, each component of Eq.~(\ref{eq:obs_causal_operator}) acquires orthogonal factors that cancel upon substitution, yielding $\mathbf{C}^{\ast}_{\tilde{\mathbf{x}}}(\mathbf{z}^{\ast}) = \mathbf{C}_{\tilde{\mathbf{x}}}(\mathbf{z})$.
The full derivation is given in Appendix~\ref{app:identifiability-proofs}.
\hfill$\blacksquare$

The theorem guarantees that the learned connectivity in observation space is unique.
Different training runs may converge to different latent representations, but as long as they differ only by an isometry, they produce the same $\mathbf{C}_{\tilde{\mathbf{x}}}$.
This removes latent-coordinate ambiguity from the reported connectivity and ensures that $\mathbf{C}_{\tilde{\mathbf{x}}}$ can be meaningfully compared across seeds, subjects, and models.
In practice, the true intrinsic dimension is unknown, and the identifiability guarantee holds when the model latent dimension $D_z$ matches the true dimension and the decoder Jacobian has full column rank.
Experimental results confirm that the learned operator reliably recovers signed coupling and its network-level organization across seeds and subjects.

\subsection{Model Architecture}

The encoder $\mathbf{m}_{\phi}$ is a multilayer perceptron (MLP) with $L$ hidden layers of width $H$ and $\tanh$ activations that maps each observation to a deterministic latent point estimate:
\begin{equation*}
\hat{\mathbf{z}}_t = \mathbf{m}_{\phi}(\mathbf{x}_t)
\end{equation*}
All network weights are initialized with the PyTorch default Kaiming uniform initialization~\cite{he2015delving}.
Using a point estimate avoids the distributional ambiguity introduced by a variational posterior; under the assumptions of Theorem~\ref{thm:identifiability}, the remaining ambiguity in the latent representation is an isometric transformation, which leaves $\mathbf{C}_{\tilde{\mathbf{x}}}$ invariant.

The drift $\boldsymbol{\mu}_{\psi}:\R^{D_z}\to\R^{D_z}$ is parameterized by an MLP of the same architecture.
Given a latent state $\hat{\mathbf{z}}_t$, the one-step Euler prediction is:
\begin{equation*}
\tilde{\mathbf{z}}_{t+\Delta t} = \hat{\mathbf{z}}_t + \Delta t\,\boldsymbol{\mu}_{\psi}(\hat{\mathbf{z}}_t)
\end{equation*}
where $\Delta t$ is set to the sampling interval of the data.

The decoder $\mathbf{f}_{\theta}$ maps latent states back to observation space through an MLP of the same architecture.

All three identifiability assumptions of Theorem~\ref{thm:identifiability} are satisfied by design.
The use of smooth $\tanh$ activations ensures that both $\boldsymbol{\mu}_\psi$ and $\mathbf{f}_\theta$ are globally Lipschitz and that $\mathbf{f}_\theta$ is twice continuously differentiable; together with the isotropic diffusion, these satisfy (A1).
The same $\tanh$ architecture also satisfies the $C^2$ requirement of (A2), and setting $D_z \le D_x$ ensures that $J_{\mathbf{f}}(\mathbf{z})$ can have full column rank.
The latent dimension and isotropic diffusion structure are fixed across runs (A3), so the architectural choices are consistent with the assumptions of Theorem~\ref{thm:identifiability}.

After training, the learned local operator is evaluated at the encoded latent state $\hat{\mathbf{z}}_t$ to obtain the estimated causal influence in observation space:
\begin{equation*}
\hat{\mathbf{C}}_t =
\Big(
J_{\mathbf{f}}(\hat{\mathbf{z}}_t)\,
J_{\boldsymbol{\mu}}(\hat{\mathbf{z}}_t)
+
(\nabla^2 \mathbf{f})(\hat{\mathbf{z}}_t)
[\boldsymbol{\mu}(\hat{\mathbf{z}}_t)]
\Big)
J_{\mathbf{f}}(\hat{\mathbf{z}}_t)^{+}
\end{equation*}
Each $\hat{\mathbf{C}}_t$ is evaluated at the time-varying latent state $\hat{\mathbf{z}}_t$, so the operator reflects the temporal dynamics of the latent trajectory.
No sliding window or temporal smoothing is used.
The resulting sequence $\{\hat{\mathbf{C}}_t\}$ forms the dEC analyzed in our experiments.

\begin{algorithm}[t]
\caption{Training of \model}
\label{alg:training}
\begin{algorithmic}[1]
\renewcommand{\algorithmicrequire}{\textbf{Input:}}
\renewcommand{\algorithmicensure}{\textbf{Output:}}
\REQUIRE Observation sequence $\{\mathbf{x}_t\}_{t=1}^{T}$, latent dimension $D_z$, sampling interval $\Delta t$, loss weights $\lambda_{\mathrm{dyn}}$ and $\lambda_{\mathrm{pred}}$.
\ENSURE Dynamic causal influence sequence $\{\hat{\mathbf{C}}_t\}$.
\STATE Initialize the encoder $\mathbf{m}_{\phi}$, the drift network $\boldsymbol{\mu}_{\psi}$, and the decoder $\mathbf{f}_{\theta}$ with Kaiming uniform initialization.
\FOR{each training epoch}
    \FOR{each adjacent observation pair $(\mathbf{x}_t, \mathbf{x}_{t+\Delta t})$}
        \STATE Encode the latent estimates $\hat{\mathbf{z}}_t$ and $\hat{\mathbf{z}}_{t+\Delta t}$.
        \STATE Predict the one-step latent state $\tilde{\mathbf{z}}_{t+\Delta t}$.
        \STATE Compute $\mathcal{L}_{\mathrm{rec}}$, $\mathcal{L}_{\mathrm{dyn}}$, and $\mathcal{L}_{\mathrm{pred}}$.
    \ENDFOR
    \STATE Update $\phi$, $\psi$, and $\theta$ by backpropagation.
\ENDFOR
\STATE Evaluate the observation-space causal operator $\hat{\mathbf{C}}_t$ at each encoded latent state $\hat{\mathbf{z}}_t$.
\RETURN the sequence $\{\hat{\mathbf{C}}_t\}$.
\end{algorithmic}
\end{algorithm}

\subsection{Objective Function}
\label{sec:objective}

We train the model with an objective defined on adjacent observations $(\mathbf{x}_t,\mathbf{x}_{t+\Delta t})$.
Consistent with the pointwise operator formulation above, we use the deterministic latent estimates:
\begin{equation*}
\hat{\mathbf{z}}_t = \mathbf{m}_{\phi}(\mathbf{x}_t),
\qquad
\hat{\mathbf{z}}_{t+\Delta t} = \mathbf{m}_{\phi}(\mathbf{x}_{t+\Delta t})
\end{equation*}
The reconstruction loss is defined as
\begin{equation*}
\mathcal{L}_{\mathrm{rec}} =
\|\mathbf{f}_{\theta}(\hat{\mathbf{z}}_t)-\mathbf{x}_t\|^2
+
\|\mathbf{f}_{\theta}(\hat{\mathbf{z}}_{t+\Delta t})-\mathbf{x}_{t+\Delta t}\|^2
\end{equation*}
Under Gaussian observation noise, minimizing $\mathcal{L}_{\mathrm{rec}}$ is equivalent to maximizing the observation likelihood.

The dynamics consistency loss aligns the encoded latent state at $t+\Delta t$ with the one-step drift prediction:
\begin{equation*}
\mathcal{L}_{\mathrm{dyn}} =
\left\|
\hat{\mathbf{z}}_{t+\Delta t}
-
\tilde{\mathbf{z}}_{t+\Delta t}
\right\|^2
\end{equation*}
Under unit isotropic diffusion, minimizing $\mathcal{L}_{\mathrm{dyn}}$ is equivalent to minimizing the negative log Euler--Maruyama transition likelihood up to scaling and an additive constant.

A prediction loss in observation space keeps the decoded drift prediction close to the next observation:
\begin{equation*}
\mathcal{L}_{\mathrm{pred}} =
\left\|
\mathbf{f}_{\theta}\!\left(
\tilde{\mathbf{z}}_{t+\Delta t}
\right)
-\mathbf{x}_{t+\Delta t}
\right\|^2
\end{equation*}

While $\mathcal{L}_{\mathrm{rec}}$ and $\mathcal{L}_{\mathrm{dyn}}$ constrain the decoder and the drift separately, $\mathcal{L}_{\mathrm{pred}}$ couples the two by requiring the decoded drift prediction to remain faithful to the observed temporal evolution.

The base training objective combines reconstruction, dynamics, and prediction:
\begin{equation*}
\mathcal{L}_{\mathrm{base}}
=
\mathcal{L}_{\mathrm{rec}}
+
\lambda_{\mathrm{dyn}}\mathcal{L}_{\mathrm{dyn}}
+
\lambda_{\mathrm{pred}}\mathcal{L}_{\mathrm{pred}}
\end{equation*}
The complete training procedure is summarized in Algorithm~\ref{alg:training}, and all network parameters are updated jointly by backpropagation on the training objective.


\section{Experiments}
\label{sec:experiments}

We evaluate the proposed framework on Lorenz96, NetSim, and HCP task fMRI datasets.
All experiments were run on a Linux workstation with an Intel Xeon Silver 4310 processor, 251~GB of RAM, and an NVIDIA A5000 GPU.
All models are optimized with AdamW under a cosine learning-rate schedule, with early stopping applied when the loss plateaus.
Hyperparameters are selected by Optuna and reported per subsection.

\subsection{Synthetic Dataset: Lorenz96}

Lorenz96~\cite{lorenz1996predictability} is a coupled chaotic system originally proposed for atmospheric modeling, defined by $\dot{x}_i = (x_{i+1} - x_{i-2})\,x_{i-1} - x_i + F$ with periodic boundary conditions.
Its chaotic dynamics and known coupling structure make it a widely used benchmark for evaluating causal learning methods.
We follow the identity observation setting of~\cite{wang2026idecdct}, where the observed signals equal the latent states, so the ground-truth causal matrix is known exactly.
The forcing coefficient is $F = 10$ and the system dimension is $N = 20$.
Each experiment generates $1000$ time steps per subject, with $4$ training and $2$ held-out test subjects; all metrics are computed on the test subjects and averaged over $5$ random seeds.
The model uses $L{=}2$ hidden layers of $H{=}512$ units, batch size 256, 2000 epochs, learning rate $1\times10^{-3}$, and loss weights $\lambda_{\mathrm{dyn}}{=}1.0$, $\lambda_{\mathrm{pred}}{=}1.0$.

Following Wang et al.~\cite{wang2026idecdct}, we report area under the receiver operating characteristic curve (AUROC), structural Hamming distance (SHD), and a cosine-similarity-based edge-alignment score, denoted WCS, which is computed on weighted directed causal matrices and captures both edge sign and magnitude.
Given the estimated causal matrix $\hat{\mathbf C}$ and the ground-truth weighted directed matrix $\mathbf C$, where the ground truth is the Jacobian of the Lorenz96 dynamics and both matrices are averaged over the evaluation time points, we compute their cosine similarity:
\begin{equation*}
\mathrm{WCS}(\hat{\mathbf C}, \mathbf C)
=
\frac{\langle \mathrm{vec}(\hat{\mathbf C}), \mathrm{vec}(\mathbf C)\rangle}
{\|\mathrm{vec}(\hat{\mathbf C})\|_2 \, \|\mathrm{vec}(\mathbf C)\|_2 }
\end{equation*}

\subsubsection{Loss Ablation}
\label{sec:loss_ablation}

\begin{figure}[t]
\centering
\includegraphics[width=\linewidth]{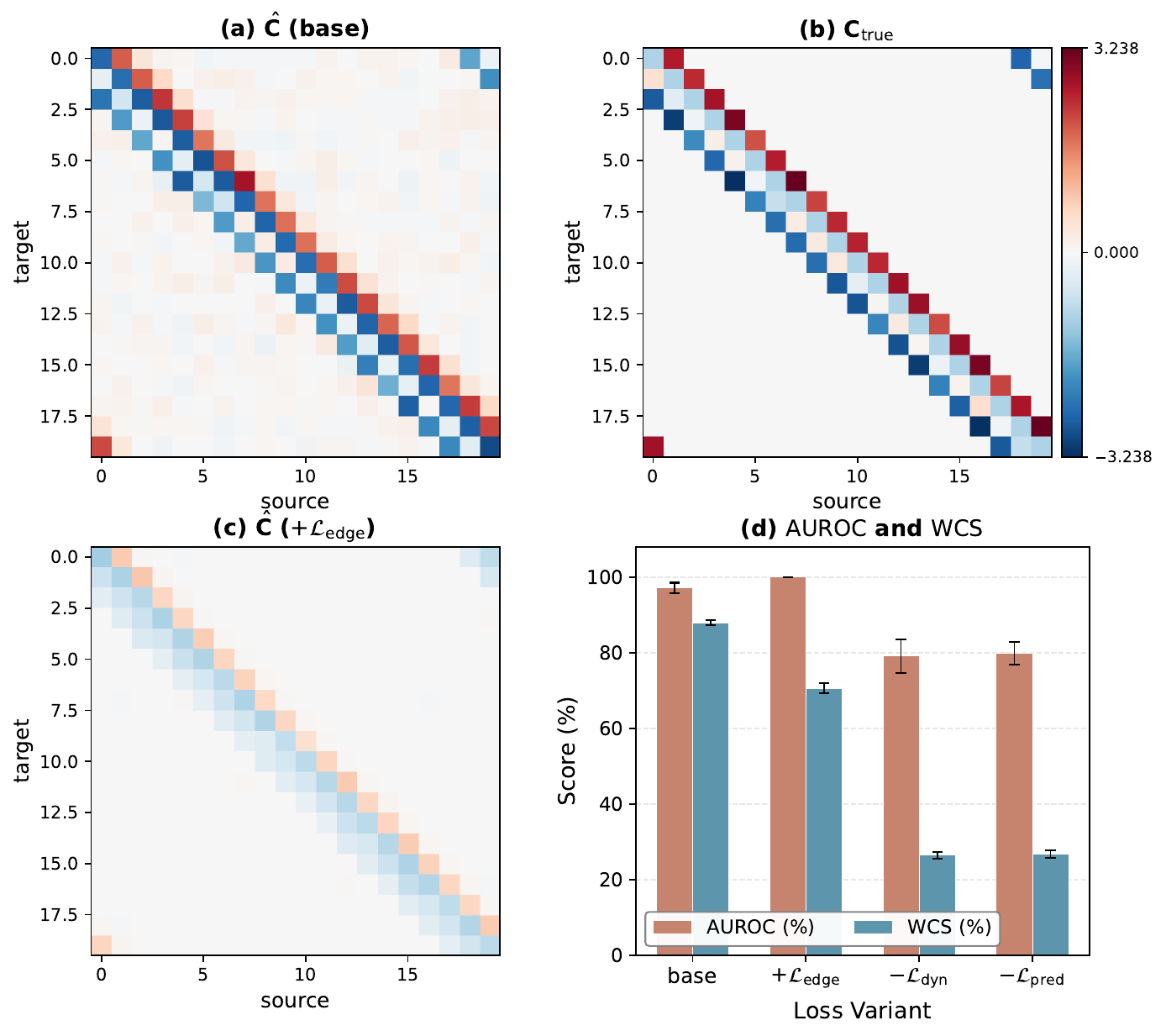}
\caption{Loss ablation and $\mathbf{C}$-matrix comparison on the Lorenz96 setting.
(a) Estimated $\hat{\mathbf{C}}$ from the base configuration.
(b) Ground-truth $\mathbf{C}_{\mathrm{true}}$ with colorbar.
(c) Estimated $\hat{\mathbf{C}}$ with $\mathcal{L}_{\mathrm{edge}}$.
(d) $\mathrm{AUROC}$ and $\mathrm{WCS}$ for the base, base augmented with $\mathcal{L}_{\mathrm{edge}}$, and ablations that remove $\mathcal{L}_{\mathrm{dyn}}$ or $\mathcal{L}_{\mathrm{pred}}$.}
\label{fig:brain26_loss_ablation}
\end{figure}

To determine whether the edge-sparsity penalty aids recovery when the ground truth is signed, we compare \model with and without $\mathcal{L}_{\mathrm{edge}}$ on the identity $20$-node Lorenz96 setting.
The edge-sparsity penalty is defined as $\mathcal{L}_{\mathrm{edge}}=\sum_{i\neq j}\sqrt{\frac{1}{A}\sum_{a}(J_{\boldsymbol{\mu}}(\hat{\mathbf{z}}^{(a)})_{i,j})^{2}+\varepsilon}$, appended to the base objective with weight $\lambda_{\mathrm{edge}}$.
Fig.~\ref{fig:brain26_loss_ablation} shows that the edge-sparsity penalty improves the binary structure score while reducing the weighted cosine alignment, as it suppresses weak edges but also shrinks true magnitudes.
The four variants are the base configuration, the base configuration augmented with $\mathcal{L}_{\mathrm{edge}}$, and two ablations that remove $\mathcal{L}_{\mathrm{dyn}}$ or $\mathcal{L}_{\mathrm{pred}}$.
Removing either term drops both the binary structure score and the weighted cosine alignment below the base configuration by a large margin, confirming that both loss terms are individually necessary.
\model keeps all three base losses and no edge penalty, and is used throughout the paper.

\subsubsection{Recovery of Causal Coupling}
\label{sec:lorenz96_identifiability}
\begin{figure}[t]
\centering
\includegraphics[width=\linewidth]{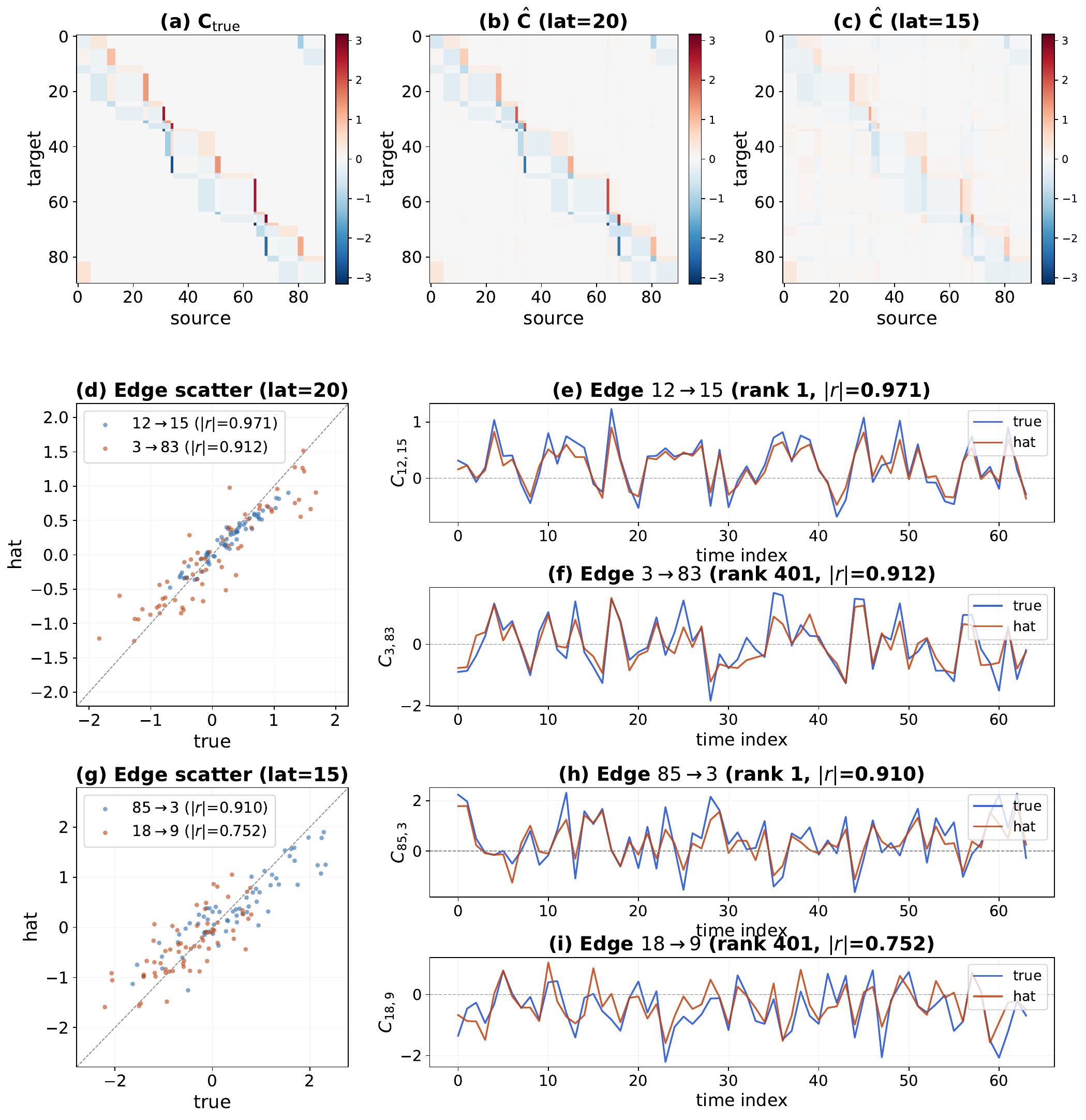}
\caption{Dynamic causal coupling tracking on the non-identity Lorenz96 setting. (a) Ground-truth $\mathbf{C}_{\mathrm{true}}$; estimated $\hat{\mathbf{C}}$ with model $D_z=20$ (b) and $D_z=15$ (c). (d--i) For $D_z=20$ and $D_z=15$, per-edge scatter of estimated versus true coupling and the time series of the top and mid-ranked edge.}
\label{fig:brain26_dynamic_edge}
\end{figure}

The continuous-time formulation tracks dynamically varying causal coupling even when the model latent dimension differs from the true intrinsic dimension.
Fig.~\ref{fig:brain26_dynamic_edge} evaluates this on the non-identity Lorenz96 setting, where the $N{=}20$ system is observed through a nonlinear map to $D_x{=}90$ dimensions and the true latent dimension is $D_z^*{=}20$.
With the model latent dimension set to $D_z=20$, the estimated $\hat{\mathbf C}$ recovers the ground-truth mean structure (Fig.~\ref{fig:brain26_dynamic_edge}a,b), and the per-edge scatter and time series confirm that the estimated coupling tracks the ground-truth temporal fluctuations for both the top-ranked and mid-ranked edges (Fig.~\ref{fig:brain26_dynamic_edge}d--f).
With a smaller model $D_z=15$, the mean structure is still recovered (Fig.~\ref{fig:brain26_dynamic_edge}c), and the dominant edge remains well tracked, while mid-ranked edges show reduced temporal fidelity (Fig.~\ref{fig:brain26_dynamic_edge}g--i).
The displayed edges are selected by within-$D_z$ correlation rank.
Dominant coupling is thus recovered reliably even under reduced latent capacity.

\subsubsection{Irregular Sampling}
\label{sec:irregular_sampling}

To assess robustness to irregularly sampled time series, we sweep $D_z \in \{10, 15, 20\}$ and dropout $\in \{20\%, 40\%, 60\%\}$ on the 20-node Lorenz96 identity setting across $5$ seeds.
Dropout randomly removes a fraction of time points while preserving the first and last observations.

\begin{table}[t]
\centering
\caption{Latent dimension and dropout grid on the 20-node identity Lorenz96 setting across $5$ seeds. Mean $\pm$ 95\% CI computed by Student's $t$-distribution.}
\label{tab:brain26_latent_dropout_grid}
\begin{tabular}{@{}lcccc@{}}
\toprule
$D_z$ & Dropout & AUROC(\%) $\uparrow$ & SHD $\downarrow$ & WCS(\%) $\uparrow$ \\
\midrule
\multirow{3}{*}{10} & $20\%$ & $71.25 \pm 1.39$ & $66.4 \pm 4.4$ & $41.21 \pm 4.69$ \\
  & $40\%$ & $70.81 \pm 7.35$ & $66.4 \pm 14.0$ & $42.42 \pm 6.60$ \\
  & $60\%$ & $78.97 \pm 7.41$ & $60.4 \pm 15.4$ & $45.80 \pm 7.98$ \\
\midrule
\multirow{3}{*}{15} & $20\%$ & $95.68 \pm 1.24$ & $22.4 \pm 4.8$ & $70.39 \pm 2.43$ \\
  & $40\%$ & $95.33 \pm 1.94$ & $22.4 \pm 4.1$ & $65.78 \pm 2.76$ \\
  & $60\%$ & $89.86 \pm 3.64$ & $36.0 \pm 9.0$ & $60.08 \pm 1.68$ \\
\midrule
\multirow{3}{*}{20} & $20\%$ & $98.56 \pm 1.01$ & $18.4 \pm 4.1$ & $82.18 \pm 1.40$ \\
  & $40\%$ & $99.15 \pm 0.34$ & $15.2 \pm 4.8$ & $68.87 \pm 2.20$ \\
  & $60\%$ & $90.26 \pm 3.12$ & $47.6 \pm 8.7$ & $45.06 \pm 3.27$ \\
\bottomrule
\end{tabular}
\end{table}

Table~\ref{tab:brain26_latent_dropout_grid} shows that the model recovers meaningful causal structure across a range of latent dimensions and observation dropout rates.
Recovery improves with latent capacity, with $D_z = 15$ achieving strong results consistent with the dynamic coupling tracking in Fig.~\ref{fig:brain26_dynamic_edge} and $D_z = 20$ reaching the best overall performance.
Moderate observation sparsity benefits binary structure at $D_z = 20$, where $40\%$ dropout yields the lowest SHD and $20\%$ dropout retains the highest weighted cosine alignment.
At $60\%$ dropout, recovery weakens for $D_z \geq 15$.
Moderate dropout may act as a form of temporal regularization by reducing the redundancy among adjacent frames, while excessive dropout removes the information needed to identify the latent dynamics.

\subsection{Synthetic Benchmark: NetSim}
NetSim~\cite{smith2011network} is a biologically motivated fMRI simulation benchmark with known ground-truth connectivity, hemodynamic convolution, and observation noise.
We evaluate on the 10-node \textit{sim2} and 50-node \textit{sim4} settings, with $D_z{=}10$ and $D_z{=}50$ respectively.
For both \textit{sim2} and \textit{sim4}, the model uses $L{=}2$, $H{=}512$, batch size 256, 100 epochs, and $\lambda_{\mathrm{dyn}}{=}1.0$, $\lambda_{\mathrm{pred}}{=}1.0$.
All metrics are averaged over $5$ seeds.
Baselines include NGM~\cite{bellot2022neural}, cMLP~\cite{tank2021neural}, CUTS~\cite{cheng2023cuts}, and CUTS+~\cite{cheng2024cuts}.

\begin{table}[t]
\centering
\caption{NetSim benchmark. Mean $\pm$ 95\% CI over $5$ seeds computed by Student's $t$-distribution.}
\label{tab:brain26_netsim_benchmark}
\setlength{\tabcolsep}{3pt}
\begin{tabular}{@{}llcccc@{}}
\toprule
Setting & Method & AUROC(\%) $\uparrow$ & SHD $\downarrow$ & WCS(\%) $\uparrow$ & Time (min) \\
\midrule
\multirow{5}{*}{\textit{sim2}} & NGM & $85.13 \pm 1.43$ & $17.2 \pm 3.8$ & $-65.09 \pm 0.45$ & 86.6 \\
 & cMLP & $86.81 \pm 0.36$ & $12.0 \pm 0.0$ & $-23.85 \pm 0.06$ & 85.3 \\
 & CUTS & $97.46 \pm 0.34$ & $9.6 \pm 2.1$ & $-28.67 \pm 0.05$ & 48.6 \\
 & CUTS+ & $91.22 \pm 0.57$ & $16.4 \pm 1.1$ & $-36.63 \pm 0.14$ & 70.7 \\
 & \model & $\mathbf{98.52 \pm 0.66}$ & $\mathbf{6.8 \pm 2.8}$ & $\mathbf{94.64 \pm 0.25}$ & $\mathbf{1.0}$ \\
\midrule
\multirow{5}{*}{\textit{sim4}} & NGM & $84.57 \pm 0.43$ & $108.8 \pm 5.4$ & $-55.96 \pm 0.20$ & 89.0 \\
 & cMLP & $91.31 \pm 0.15$ & $90.4 \pm 5.4$ & $-14.62 \pm 0.04$ & 371.0 \\
 & CUTS & $\mathbf{98.65 \pm 0.16}$ & $56.8 \pm 2.8$ & $-9.41 \pm 0.01$ & 155.3 \\
 & CUTS+ & $95.98 \pm 0.41$ & $76.4 \pm 6.2$ & $-11.71 \pm 0.07$ & 192.9 \\
 & \model & $97.19 \pm 1.42$ & $\mathbf{55.6 \pm 5.4}$ & $\mathbf{92.74 \pm 0.24}$ & $\mathbf{1.1}$ \\
\bottomrule
\end{tabular}
\end{table}

Table~\ref{tab:brain26_netsim_benchmark} shows that \model achieves the best AUROC, SHD, and WCS on \textit{sim2}.
On the larger \textit{sim4} setting, CUTS attains a slightly higher AUROC, as its thresholded binary output suits the rank-based metric on \textit{sim4}'s sparse ground truth.
\model achieves the lowest SHD and produces strongly positive WCS values while all baselines show negative ones.
This sign gap is expected by construction, as standard methods output unsigned edge strengths, so their cosine similarity against a ground truth containing negative self-inhibition is necessarily negative, whereas the latent SDE Jacobian in \model naturally preserves the sign of coupling.
The WCS comparison therefore reflects the additional information captured by signed estimation rather than a difference in unsigned recovery quality.
\model is also faster than all baselines, completing each seed in approximately one minute.
\model generalizes to held-out test subjects, whereas baselines require per-subject fitting and report in-sample fit.

\begin{figure}[t]
\centering
\includegraphics[width=\linewidth]{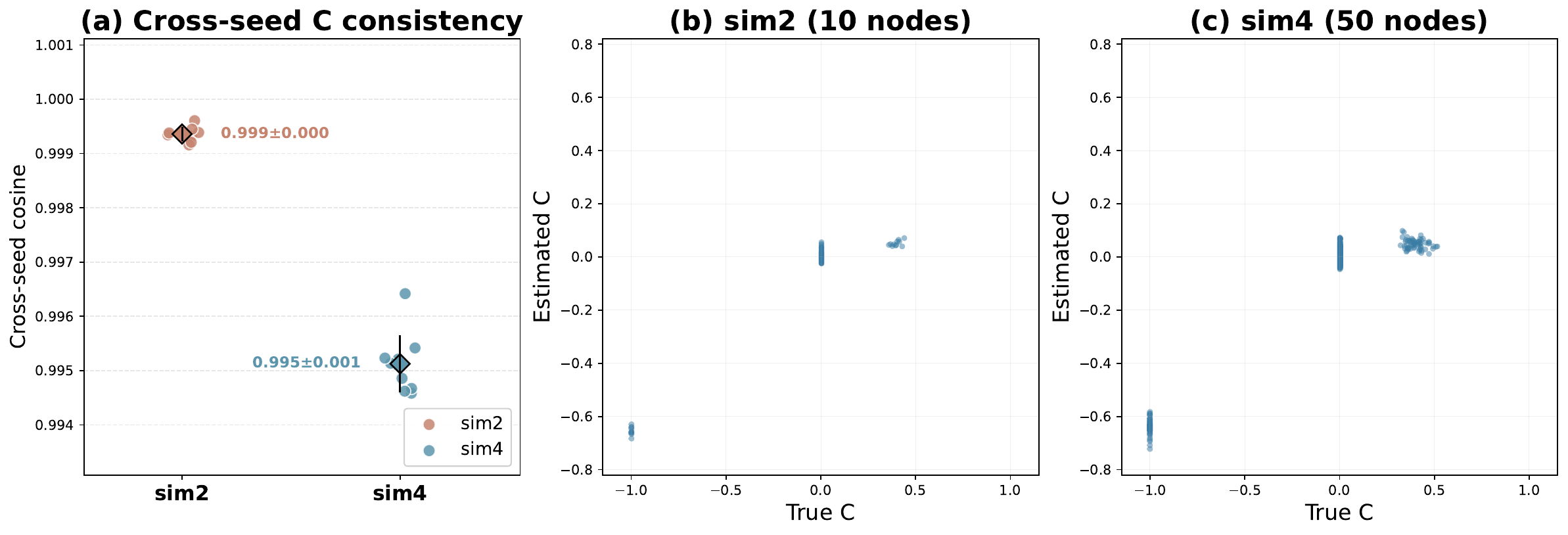}
\caption{NetSim results. (a) Cross-seed pairwise cosine similarity of $\mathbf{C}$ matrices across 5 independent seeds, upper triangle for \textit{sim2} and lower triangle for \textit{sim4}. (b)--(c) Scatter of estimated versus true edge weights.}
\label{fig:brain26_netsim_combined}
\end{figure}

Fig.~\ref{fig:brain26_netsim_combined} summarizes cross-seed consistency of the estimated causal matrix and the scatter of estimated versus true edge weights.
\model produces consistent $\mathbf{C}$ across independent training seeds on both \textit{sim2} and \textit{sim4}.
The scatter plots reveal three natural clusters determined by the NetSim ground-truth structure.
These correspond to diagonal self-inhibition at $x=-1$, absent edges at $x=0$, and a sparse set of positive connections in the first quadrant.
\model correctly recovers self-inhibition on the diagonal, assigns near-zero values to absent edges, and captures the correct sign and magnitude for true connections.

\subsection{Real fMRI Dataset: HCP Social Cognition}

We evaluate \model on real fMRI using the Human Connectome Project (HCP) Young Adult dataset~\cite{vanessen2013wumin}, a large and openly available cohort of healthy adults acquired under a harmonized high-resolution protocol.
We use the Social Cognition task-fMRI protocol~\cite{barch2013function}, which employs a theory-of-mind paradigm.
The sample follows the official HCP unrelated-subjects release and comprises 339 subjects with no known family structure.
Each subject contributes the left-to-right phase-encoding run.

The Social Cognition task presents mentalizing and random interaction conditions as interleaved blocks~\cite{barch2013function} within a single left-to-right phase-encoding run of 274 BOLD frames, of which 99 are mentalizing and 66 are random.
For each subject, \model produces a per-time-point causal coupling matrix.

Functional runs were acquired on a 3T scanner with a 720 ms repetition time, a 33.1 ms echo time, 2.0 mm isotropic voxels, and a multiband factor of eight.
The volumetric HCP minimally preprocessed release~\cite{glasser2013minimal} was used as input, where gradient distortion, susceptibility distortion, head motion, and registration to standard space had already been corrected.
Region-level time series were then obtained by resampling the Schaefer atlas~\cite{schaefer2018local} onto the functional grid and averaging the voxels within each of the 200 cortical parcels.
No additional confound regression or temporal filtering was applied, so the task-evoked structure is retained in the input.
All 339 subjects are used for both model training and the analyses, so the results are in-sample descriptive findings.
The seven-network partition of the atlas assigns each parcel to one of the canonical resting-state networks (RSNs): visual (VIS), somatomotor (SOM), dorsal-attention (DAN), ventral-attention (VAN), limbic (LIM), frontoparietal (FPN), and default-mode (DMN).

The model uses $L{=}2$ hidden layers of $H{=}512$ units, batch size 256, 200 epochs, learning rate $1\times10^{-3}$ under a cosine schedule, and loss weights $\lambda_{\mathrm{dyn}}=1.0$, $\lambda_{\mathrm{pred}}=1.0$.
A subject context encoder allows a single shared model to adapt to individual subjects.
All results are averaged over five independent seeds.

\subsubsection{Seed Stability of Effective Connectivity}
\label{sec:invariant_stability}

To assess whether the learned connectivity representation is robust to the randomness of model initialization, we trained \model with five independent seeds on the same HCP Social Cognition dataset and compared the resulting connectivity matrices at the region-of-interest (ROI) and RSN levels, both signed and unsigned.
All cross-seed comparisons in this subsection apply per-subject full-matrix normalization.
Each subject's connectivity matrix is divided by its total absolute mass before averaging, removing inter-subject differences in global coupling magnitude:
\begin{equation}
\tilde{\mathbf C}_{\text{subj}} = \frac{\hat{\mathbf C}_{\text{subj}}}{\sum_{i,j} |\hat{\mathbf C}_{\text{subj},ij}|},\quad |\tilde{\mathbf C}_{\text{subj}}| = \frac{|\hat{\mathbf C}_{\text{subj}}|}{\sum_{i,j} |\hat{\mathbf C}_{\text{subj},ij}|}
\label{eq:per_subj_norm}
\end{equation}

Table~\ref{tab:brain26_latent_sweep} reports the cross-seed consistency at $D_z{=}15$ and $D_z{=}20$.
At the RSN level, unsigned coupling is nearly identical across seeds at both dimensions, and the RSN-level consistency is higher than at the ROI level, as aggregation over regions within each network averages out ROI-level estimation noise.
At the ROI level, unsigned magnitude is consistent across seeds and decreases modestly as the latent dimension increases.
Signed coupling is also consistent across seeds after normalization, with the RSN-level signed cosine approaching the corresponding unsigned values at $D_z{=}15$, supporting network-level analysis of the sign pattern.
The subsequent RSN-level analyses are reported at $D_z{=}15$.

\begin{table}[ht]
\centering
\caption{Cross-seed consistency of \model on HCP Social Cognition. Mean $\pm$ 95\% CI over $\binom{5}{2}=10$ pairwise seed comparisons, computed by bootstrap resampling of the pairwise cosine.}
\label{tab:brain26_latent_sweep}
\begin{tabular}{l c c}
\toprule
Metric & $D_z{=}15$ & $D_z{=}20$ \\
\midrule
RSN $|\tilde{\mathbf C}|$ cosine & $0.986 \pm 0.006$ & $0.989 \pm 0.005$ \\
ROI $|\tilde{\mathbf C}|$ cosine & $0.950 \pm 0.015$ & $0.932 \pm 0.015$ \\
RSN signed $\tilde{\mathbf C}$ cosine & $0.935 \pm 0.017$ & $0.915 \pm 0.030$ \\
ROI signed $\tilde{\mathbf C}$ cosine & $0.914 \pm 0.019$ & $0.858 \pm 0.041$ \\
\bottomrule
\end{tabular}
\end{table}

\subsubsection{ROI- and RSN-Level Organization}
\label{sec:rsn_organization}
\begin{figure}[t]
\centering
\includegraphics[width=\linewidth]{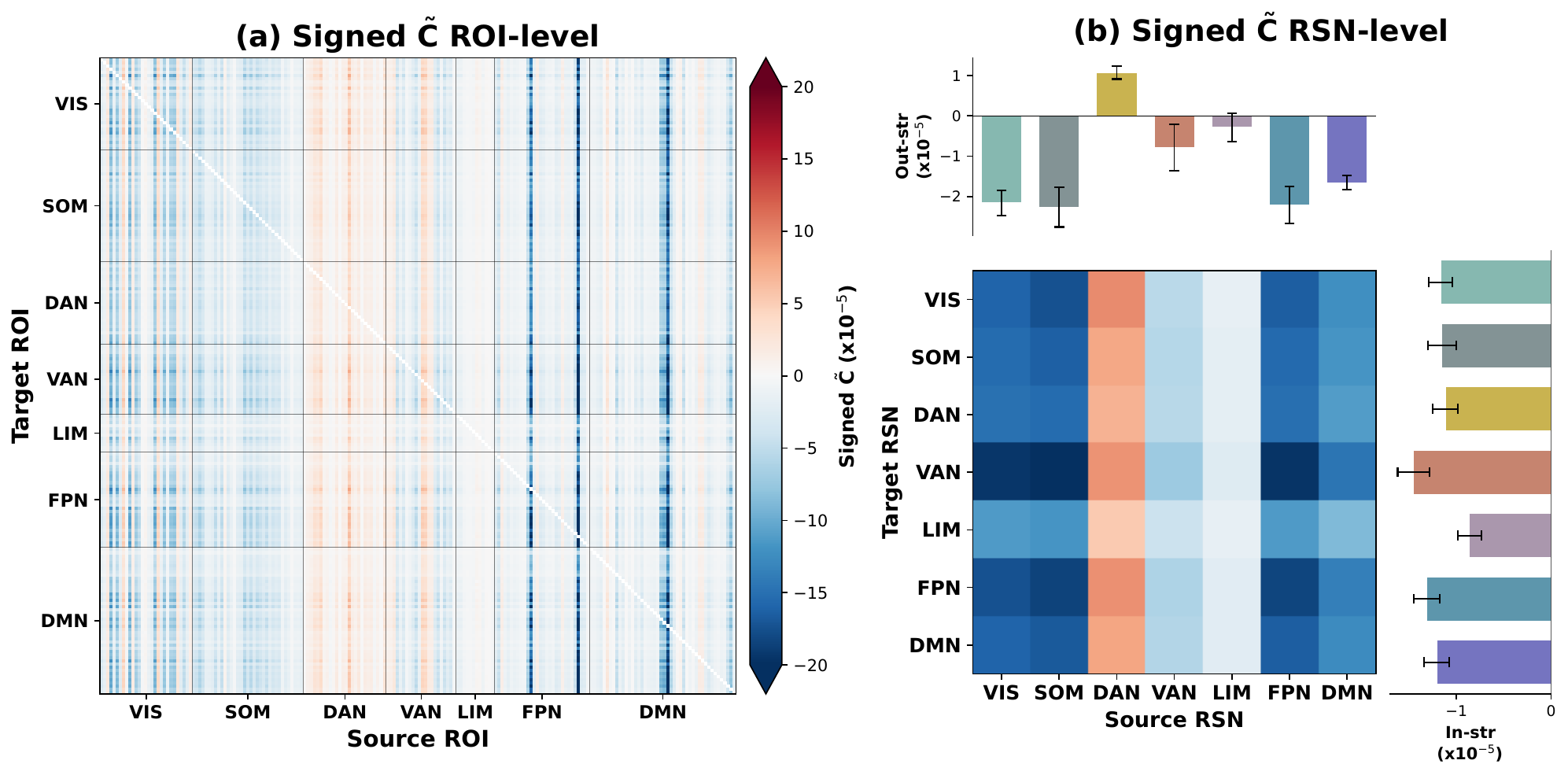}
\caption{
Signed dEC architecture.
(a) Per-subject normalized $200\times200$ ROI-level $\tilde{\mathbf C}$ matrix at $D_z=15$.
(b) Corresponding $7\times7$ RSN-aggregated signed matrix with marginal bar charts showing per-RSN out-strength and in-strength; error bars are cross-seed standard deviation.
}
\label{fig:brain26_rsn_signed}
\end{figure}

To characterize the EC architecture, Fig.~\ref{fig:brain26_rsn_signed}a shows the per-subject normalized $200\times200$ ROI-level signed $\tilde{\mathbf C}$ matrix.
The most salient feature is a set of vertical blue bands, each corresponding to a source ROI that exerts global inhibitory observation-level influence across the cortex; the strongest such bands originate from the bilateral control-cingulate parcels in FPN and the posterior-cingulate parcels in DMN.
The two strongest inhibitory sources at the ROI level are both cingulate-region parcels, the control-cingulate parcels of FPN and the posterior-cingulate parcels of DMN.
The posterior cingulate cortex is a core node of the default-mode network~\cite{spreng2009common} and the control-cingulate parcels form the cingulate segment of the frontoparietal control network~\cite{smith2009correspondence}, indicating that observation-level inhibition over distributed cortex is routed through the cingulate region.

Aggregating into $7\times7$ directed RSN matrices (Fig.~\ref{fig:brain26_rsn_signed}b), the DAN column is entirely positive while all other columns are negative, reflecting that DAN is the only consistently positive out-strength source.
The out-strength and in-strength bar charts (Fig.~\ref{fig:brain26_rsn_signed}b) confirm this difference.
All networks receive net inhibitory input, with VAN and FPN receiving the strongest inhibition, while SOM, FPN, and VIS are the strongest inhibitory sources.
LIM has the weakest magnitude in both directions, indicating that it participates minimally in the directed coupling during social cognition.
The positive observation-level out-strength of DAN is consistent with its established function in top-down attentional control~\cite{corbetta2002control}, and this excitatory outflow at the observation level may reflect top-down attentional enhancement overriding the suppressive influence of feedback predictions in predictive coding~\cite{bastos2012canonical}.
The positive influence from DAN to DMN differs from the resting-state anticorrelation between these networks~\cite{fox2005anticorrelated}, as DAN is recruited for attention to external social stimuli and DMN for theory-of-mind processing~\cite{spreng2009common}.
The negative observation-level outflow from FPN and DMN is consistent with their roles in cognitive control and internally directed thought, respectively~\cite{smith2009correspondence}.

\subsubsection{Condition-Specific Effective Connectivity}
\label{sec:condition_specific}

\begin{figure}[htbp]
\centering
\includegraphics[width=\linewidth]{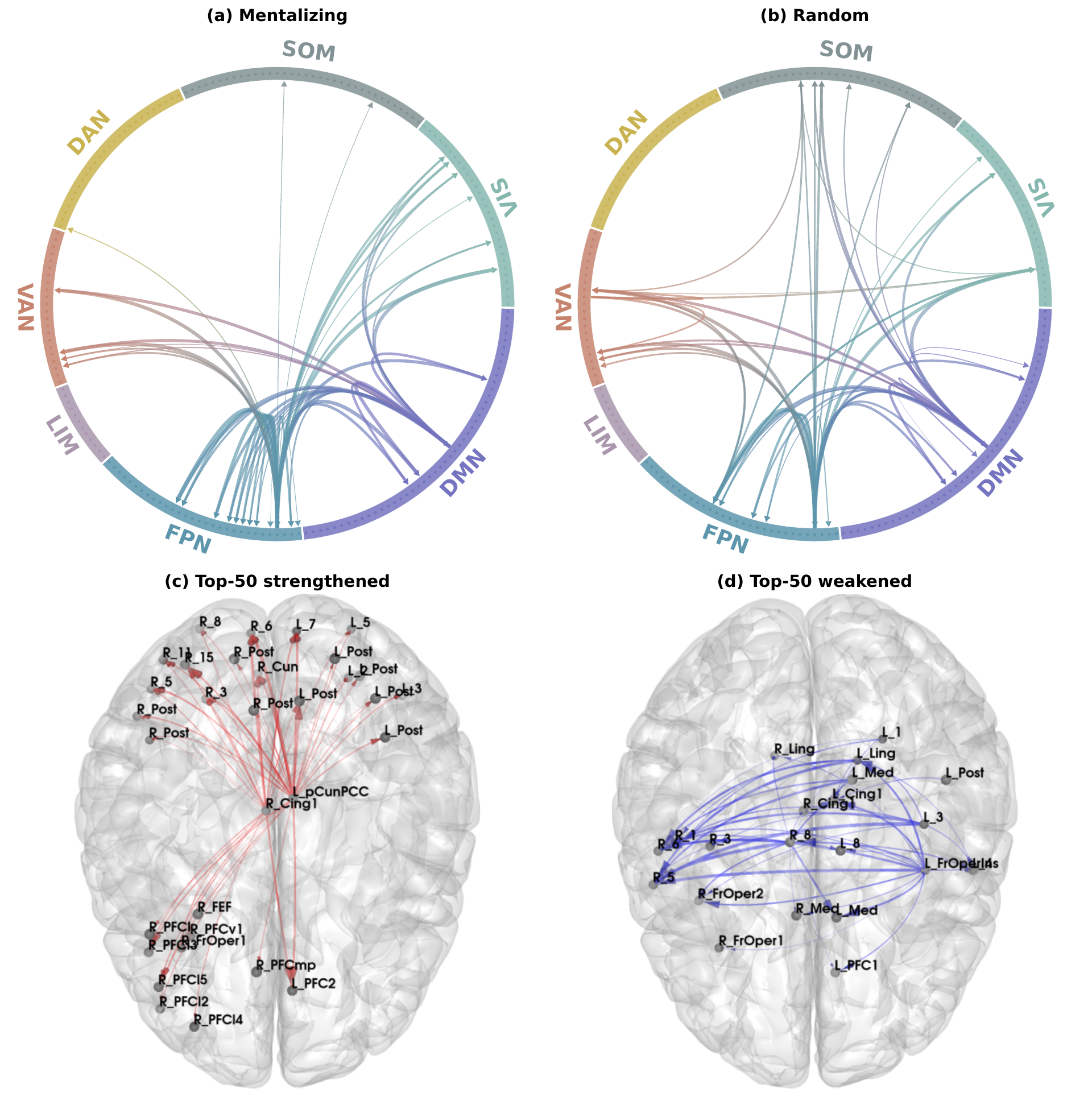}
\caption{Condition-specific dEC at $D_z=15$. (a)~Mentalizing condition circos of the top-50 strongest ROI-level edges; (b)~random condition circos. Edge width encodes connection magnitude. (c)~Dorsal view of the 50 strongest mentalizing-strengthened directed edges ranked by $|\Delta|\tilde{\mathbf C}||$; (d)~dorsal view of the 50 strongest mentalizing-weakened directed edges.}
\label{fig:brain26_circos_dims}
\end{figure}

To assess whether the recovered EC encodes cognitive state in addition to subject identity, separate models were trained exclusively on mentalizing or random time points at $D_z=15$.
Training per condition instead of comparing time windows within a single pooled model is designed to attribute observed EC differences to the cognitive state rather than to averaging over heterogeneous dynamics.
Mentalizing and random frames were identified from the HCP task event design and extracted as separate subsequences, discarding all frames outside the target condition.
Because the two conditions are interleaved within a single run, the extracted subsequences differ in their temporal continuity and may carry residual hemodynamic transition effects at block boundaries.
Five seeds were trained per condition on 339 subjects.
The strongest directed edges from each condition are shown in Fig.~\ref{fig:brain26_circos_dims}.
The condition difference is defined as $\Delta|\tilde{\mathbf C}| = |\tilde{\mathbf C}_{\mathrm{mental}}| - |\tilde{\mathbf C}_{\mathrm{random}}|$ using per-subject normalization (Eq.~\ref{eq:per_subj_norm}).
Both conditions share a source backbone in FPN and DMN, with the VAN and FPN as the primary targets.
Differences are confined to edge magnitudes, without changes in the presence or absence of specific pathways.
The brainnet dorsal views in Fig.~\ref{fig:brain26_circos_dims}c and~d separate the mentalizing-strengthened and mentalizing-weakened edges.
The 50 strengthened edges are all driven by FPN and DMN outflow, while the 50 weakened edges involve VAN and SOM as targets.
The strengthened edges are driven by only two source parcels, the left posterior-cingulate in DMN and the right control-cingulate in FPN, which also account for the entire mentalizing intensity top-50 (Fig.~\ref{fig:brain26_circos_dims}a).
The weakened edges instead spread across eight sources in VAN and SOM and are received mainly by SOM and VAN.

\begin{figure}[htbp]
\centering
\includegraphics[width=\linewidth]{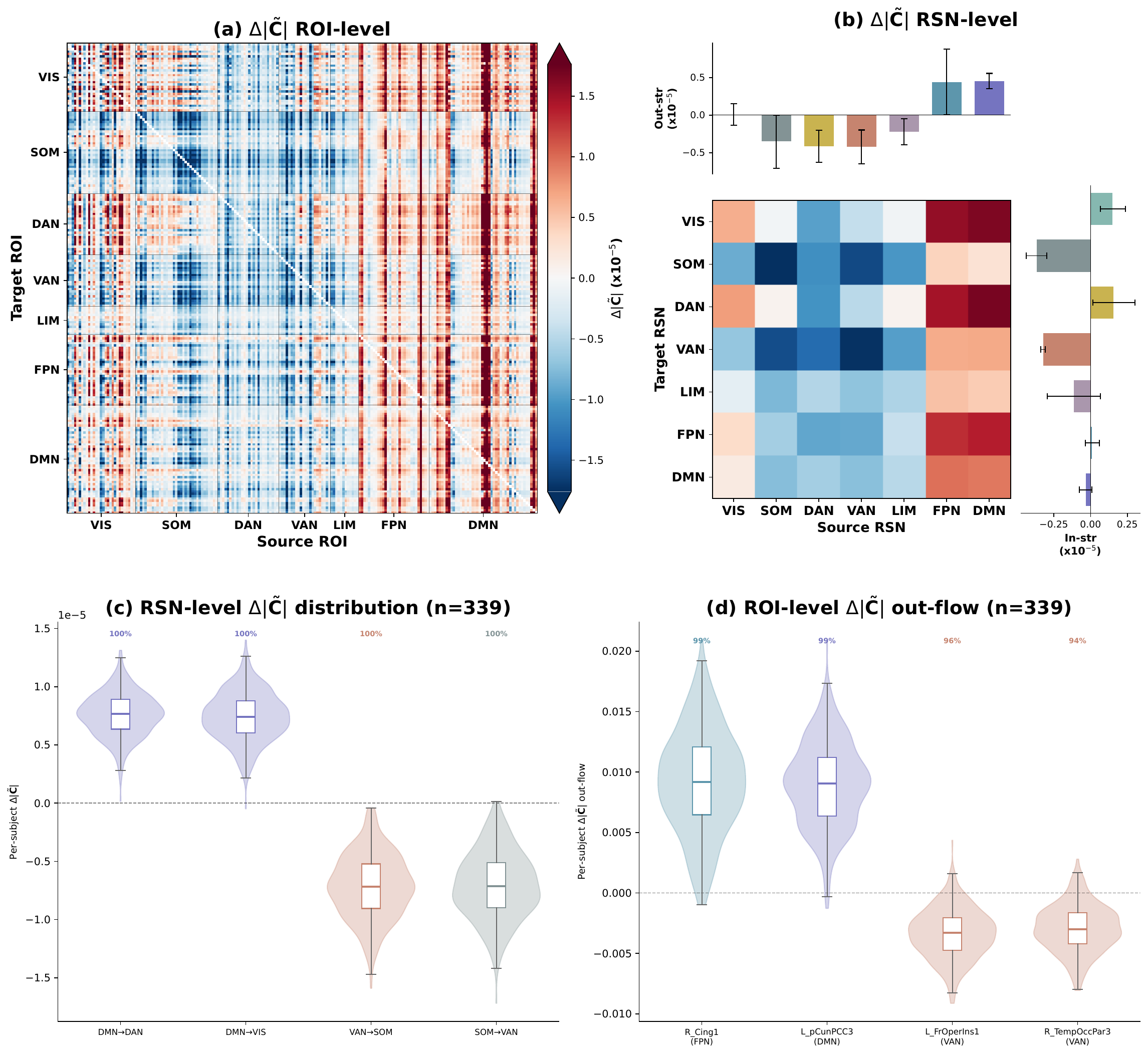}
\caption{Condition-specific effective connectivity difference. (a)~$\Delta|\tilde{\mathbf C}|$ at the ROI level, subject-average across five seeds. (b)~RSN-level $\Delta|\tilde{\mathbf C}|$ with marginal bars showing per-RSN out-strength and in-strength; error bars are cross-seed standard deviation. (c-d)~Per-subject distribution (n=339) of directed RSN-level and ROI-level $\Delta|\tilde{\mathbf C}|$ for the top-4 strongest RSN pairs and the top-4 hub ROIs.}
\label{fig:brain26_delta_c_rsn}
\end{figure}

Fig.~\ref{fig:brain26_delta_c_rsn}a shows the ROI-level $\Delta|\tilde{\mathbf C}|$ heatmap, where the largest strengthened differences originate from FPN and DMN networks.
The RSN-level heatmap in Fig.~\ref{fig:brain26_delta_c_rsn}b confirms that positive incoming differences concentrate on DAN and VIS as targets, while DMN shows negative incoming differences.
Per-subject distributions confirm that the RSN-level pattern holds across individuals, with the strongest differences sharing a common sign in nearly all subjects (Fig.~\ref{fig:brain26_delta_c_rsn}c), and the top hub ROIs show consistent sign agreement at the ROI level (Fig.~\ref{fig:brain26_delta_c_rsn}d).
Negative differences concentrate on VAN and SOM.
The positive incoming difference to DAN is consistent with its involvement in top-down attentional deployment during social cognition~\cite{corbetta2002control}.
The negative incoming difference of DMN is consistent with its role as the core substrate of theory-of-mind processing~\cite{spreng2009common}, reflecting reduced incoming influence during mentalizing as internally directed processing dominates over externally driven input.
Coupling strength is thus redirected from DMN toward attention and visual networks during mentalizing.

\begin{table}[htbp]
\centering
\caption{Condition-specific EC stability and difference metrics.}
\label{tab:condition_stability}
\begin{tabular}{lc}
\toprule
Metric & Value \\
\midrule
Cross-condition cosine ($|\tilde{\mathbf C}_{\mathrm{mental}}|$, $|\tilde{\mathbf C}_{\mathrm{random}}|$) & $0.929\pm0.024$ \\
Cross-seed cosine, mentalizing & $0.948\pm0.015$ \\
Cross-seed cosine, random & $0.939\pm0.025$ \\
\midrule
Top-20 RSN pair by $|\Delta|\tilde{\mathbf C}||$, 5/5 sign agreement & 16 / 20 \\
\quad strengthened / weakened & 8 / 12 \\
Top-400 ROI pair by $|\Delta|\tilde{\mathbf C}||$, 5/5 sign agreement & 349 / 400 \\
\quad strengthened / weakened & 384 / 16 \\
\quad FPN-source (5/5) & 148 / 183 \\
\quad DMN-source (5/5) & 180 / 196 \\
\bottomrule
\end{tabular}
\end{table}

Table~\ref{tab:condition_stability} summarizes the stability and difference metrics.
Within each condition, cross-seed consistency remains high, confirming that condition-specific training does not degrade the stability of the learned representation.
The cross-condition cosine remains high, consistent with the shared backbone observed in Fig.~\ref{fig:brain26_circos_dims}.
The condition difference is systematic at the RSN level.
Among the top-20 strongest RSN pairs, most achieve full sign agreement across seeds in $\Delta|\tilde{\mathbf C}|$, with a roughly even split between strengthened and weakened differences.
Among the top-400 strongest ROI pairs, most achieve full sign agreement across seeds, with strengthened differences exceeding weakened ones.
The top-400 pairs originate mainly from FPN and DMN, indicating that the largest condition differences are driven by increased causal outflow from these two networks during mentalizing.
Weakened differences are distributed across many weak ROI pairs within VAN and SOM, producing notable RSN-level averages but only 16 individual ROI pairs strong enough to enter the top-400.

The HCP analyses show that \model recovers a signed dEC architecture that is stable across seeds, organized at the network level, and reconfigures with cognitive state.
The recovered signed dEC thus provides a network-level description of brain organization and its task-dependent reconfiguration.


\section{Conclusion}
\label{sec:conclusion}

We have presented \model, a latent SDE framework that learns dEC from high-dimensional neuroimaging data through a rational, identifiable, signed causal influence operator defined in observation space.
Under the invariance assumptions of the underlying dynamics, the recovered connectivity is uniquely determined in observation space.
On synthetic benchmarks with ground truth, \model recovers the causal structure on Lorenz96 and achieves the lowest SHD among compared methods on NetSim, while its signed output separates excitatory from inhibitory observation-level influences that unsigned baselines cannot represent.
On HCP Social Cognition task-fMRI, the rational and identifiable operator recovers connectivity consistent across seeds and reveals a signed architecture aligned with the functional organization of social cognition. The signed architecture identifies the dorsal-attention network as the only consistently positive out-strength source, with the somatomotor, frontoparietal, and visual networks as the strongest inhibitory sources.
Condition-specific training further reveals EC reconfiguration across task states, with causal outflow from the default-mode and frontoparietal networks strengthened during theory-of-mind processing.


\appendices

\section{Proofs of the Main Results}
\label{app:identifiability-proofs}

\noindent\textbf{Theorem~\ref{thm:rationality} (Rationality).} Let $\dot{\tilde{\mathbf{x}}} = J_{\mathbf{f}}(\mathbf{z})\boldsymbol{\mu}(\mathbf{z})$ denote the velocity of $\tilde{\mathbf{x}}$.
Under the assumptions of Definition~\ref{def:obs_causal_operator}, for any tangent perturbation $d\mathbf{x}\in T_{\tilde{\mathbf{x}}}\mathcal{M}$, the induced local response of the observation velocity satisfies
\begin{equation*}
d\dot{\tilde{\mathbf{x}}} = \mathbf{C}_{\tilde{\mathbf{x}}}(\mathbf{z})\,d\mathbf{x}
\end{equation*}

\begin{proof}
By the chain rule, the observation velocity is $\dot{\tilde{\mathbf{x}}}=\frac{d}{dt}\mathbf{f}(\mathbf{z}(t))=J_{\mathbf{f}}(\mathbf{z})\,\dot{\mathbf{z}}=J_{\mathbf{f}}(\mathbf{z})\,\boldsymbol{\mu}(\mathbf{z})$. Taking the first-order differential of this product and applying the product rule produces two terms, the local rate of change of the latent dynamics and the curvature of the observation map:
\begin{equation*}
d\dot{\tilde{\mathbf{x}}}=\Big(J_{\mathbf{f}}(\mathbf{z})J_{\boldsymbol{\mu}}(\mathbf{z})+(\nabla^2\mathbf{f})(\mathbf{z})[\boldsymbol{\mu}(\mathbf{z})]\Big)\,d\mathbf{z}
\end{equation*}
Since $J_{\mathbf{f}}(\mathbf{z})$ has full column rank and the tangent perturbation satisfies $d\mathbf{x}\in T_{\tilde{\mathbf{x}}}\mathcal{M}=\mathrm{Im}(J_{\mathbf{f}}(\mathbf{z}))$, there exists a unique latent perturbation $d\mathbf{z}$ with $d\mathbf{x}=J_{\mathbf{f}}(\mathbf{z})\,d\mathbf{z}$, equivalently $d\mathbf{z}=J_{\mathbf{f}}(\mathbf{z})^{+}d\mathbf{x}$. Substituting gives $d\dot{\tilde{\mathbf{x}}}=\mathbf{C}_{\tilde{\mathbf{x}}}(\mathbf{z})\,d\mathbf{x}$.
\end{proof}

\begin{lemma}[Pseudoinverse orthogonal invariance~\cite{penrose1955generalized}]
\label{lem:pseudoinverse-orthogonal}
Let $A \in \R^{m \times d}$ and $Q \in \R^{d \times d}$ be orthogonal ($Q^\top Q = QQ^\top = I_d$). Then
\begin{equation*}
(AQ^\top)^+ = Q\,A^+
\end{equation*}
\end{lemma}

This identity is a standard property of the Moore--Penrose pseudoinverse~\cite{penrose1955generalized} and is used in the proof of Lemma~\ref{lem:operator-invariance}.

\begin{lemma}[Isometric invariance of $\mathbf{C}_{\tilde{\mathbf{x}}}$]
\label{lem:operator-invariance}
Let $\mathbf{f} \in C^2$ be a smooth immersion with $J_{\mathbf{f}}(\mathbf{z})$ of full column rank. Suppose two parameterizations $(\mathbf{f}, \boldsymbol{\mu})$ and $(\mathbf{f}^*, \boldsymbol{\mu}^*)$ are related by an isometric transformation $\mathbf{z}^* = Q\mathbf{z} + \mathbf{b}$ with $Q$ orthogonal, so that $\mathbf{f}^*(\mathbf{z}^*) = \mathbf{f}(\mathbf{z})$ and $\boldsymbol{\mu}^*(\mathbf{z}^*) = Q\,\boldsymbol{\mu}(\mathbf{z})$. Then, for any tangent perturbation $d\mathbf{x} \in T_{\tilde{\mathbf{x}}}\mathcal{M}$,
\begin{equation*}
\mathbf{C}^*_{\tilde{\mathbf{x}}}(\mathbf{z}^*)\,d\mathbf{x} = \mathbf{C}_{\tilde{\mathbf{x}}}(\mathbf{z})\,d\mathbf{x}
\end{equation*}
\end{lemma}

\begin{proof}
We verify the invariance by substituting the isometric relations into each component of $\mathbf{C}^*_{\tilde{\mathbf{x}}}$.

For the decoder, $\mathbf{f}^*(\mathbf{z}^*) = \mathbf{f}(\mathbf{z})$ with $\mathbf{z} = Q^\top(\mathbf{z}^* - \mathbf{b})$ gives
\begin{equation*}
J_{\mathbf{f}^*}(\mathbf{z}^*) = J_{\mathbf{f}}(\mathbf{z})\,Q^\top, \qquad
J_{\mathbf{f}^*}^+(\mathbf{z}^*) = (J_{\mathbf{f}}(\mathbf{z})\,Q^\top)^+ = Q\,J_{\mathbf{f}}(\mathbf{z})^+
\end{equation*}
where the second equality follows from Lemma~\ref{lem:pseudoinverse-orthogonal}.
For the drift, $\boldsymbol{\mu}^*(\mathbf{z}^*) = Q\,\boldsymbol{\mu}(\mathbf{z})$ gives
\begin{equation*}
J_{\boldsymbol{\mu}^*}(\mathbf{z}^*) = Q\,J_{\boldsymbol{\mu}}(\mathbf{z})\,Q^\top
\end{equation*}
For the Hessian contraction, the chain rule with $\mathbf{z} = Q^\top(\mathbf{z}^* - \mathbf{b})$ (affine, hence zero second derivative) gives, for each $i,j,k$,
\begin{equation*}
\frac{\partial^2 f_i^*(\mathbf{z}^*)}{\partial z_j^* \partial z_k^*}
= \sum_{a,b} \frac{\partial^2 f_i(\mathbf{z})}{\partial z_a \partial z_b}\,Q_{ja}\,Q_{kb}
\end{equation*}
Contracting with $\mu_k^*(\mathbf{z}^*) = \sum_c Q_{kc}\,\mu_c(\mathbf{z})$, we obtain
\begin{equation*}
\bigl[(\nabla^2 \mathbf{f}^*)\boldsymbol{\mu}^*(\mathbf{z}^*)\bigr]_{ij}
= \sum_{a,c} \frac{\partial^2 f_i(\mathbf{z})}{\partial z_a \partial z_c}\,Q_{ja}\,\mu_c(\mathbf{z})
= \bigl[\bigl((\nabla^2 \mathbf{f})\boldsymbol{\mu}(\mathbf{z})\bigr)\,Q^\top\bigr]_{ij}
\end{equation*}
Combining the three components and substituting into the definition of $\mathbf{C}^*_{\tilde{\mathbf{x}}}$ gives
\begin{align*}
\mathbf{C}^*_{\tilde{\mathbf{x}}}(\mathbf{z}^*)\,d\mathbf{x}
&= \Bigl(J_{\mathbf{f}^*}J_{\boldsymbol{\mu}^*} + (\nabla^2 \mathbf{f}^*)\boldsymbol{\mu}^*\Bigr) J_{\mathbf{f}^*}^+\,d\mathbf{x} \\
&= \Bigl((J_{\mathbf{f}}Q^\top)(Q\,J_{\boldsymbol{\mu}}Q^\top) + (\nabla^2 \mathbf{f})\boldsymbol{\mu}\,Q^\top\Bigr)(Q\,J_{\mathbf{f}}^+)\,d\mathbf{x} \\
&= \Bigl(J_{\mathbf{f}}J_{\boldsymbol{\mu}} + (\nabla^2 \mathbf{f})\boldsymbol{\mu}\Bigr)(Q^\top Q)\,J_{\mathbf{f}}^+\,d\mathbf{x} \\
&= \Bigl(J_{\mathbf{f}}J_{\boldsymbol{\mu}} + (\nabla^2 \mathbf{f})\boldsymbol{\mu}\Bigr)J_{\mathbf{f}}^+\,d\mathbf{x} \\
&= \mathbf{C}_{\tilde{\mathbf{x}}}(\mathbf{z})\,d\mathbf{x}
\end{align*}
\end{proof}

\noindent\textbf{Theorem~\ref{thm:identifiability} (Identifiability).} Assume that: (A1) the drift $\boldsymbol{\mu}$ and diffusion $\boldsymbol{\sigma}$ of the latent SDE are globally Lipschitz, $\boldsymbol{\sigma}$ is symmetric positive definite and reducible, meaning the SDE admits an invertible transform that reduces its diffusion to isotropic form, and the observation noise $\boldsymbol{\epsilon}$ is Gaussian with non-degenerate covariance; (A2) the observation map $\mathbf{f}$ is $C^2$ and $J_{\mathbf{f}}(\mathbf{z})$ has full column rank; and (A3) two parameterizations share the same latent dimension and isotropic diffusion. Let $(\mathbf{f},\boldsymbol{\mu})$ and $(\mathbf{f}^{\ast},\boldsymbol{\mu}^{\ast})$ be two parameterizations that induce the same joint distribution of adjacent observations $(\mathbf{x}_t,\mathbf{x}_{t+\Delta t})$. Then, for every common observation point $\tilde{\mathbf{x}}=\mathbf{f}(\mathbf{z})=\mathbf{f}^{\ast}(\mathbf{z}^{\ast})$,
\begin{equation*}
\mathbf{C}^{\ast}_{\tilde{\mathbf{x}}}(\mathbf{z}^{\ast}) = \mathbf{C}_{\tilde{\mathbf{x}}}(\mathbf{z})
\end{equation*}
Thus, the observation-space causal operator is identifiable in observation coordinates up to isometric latent transformations.

\begin{proof}
Under the assumptions of Theorem~\ref{thm:identifiability}, by the identifiability result of Hasan et al.~\cite{hasan2021identifying}, if two models generate the same joint distribution of adjacent observations, their latent representations differ at most by an isometric transformation, with an orthogonal matrix $Q$ and a translation $\mathbf{b}$ such that $\mathbf{z}^* = Q\mathbf{z} + \mathbf{b}$, with $\mathbf{f}^*(\mathbf{z}^*) = \mathbf{f}(\mathbf{z})$ and $\boldsymbol{\mu}^*(\mathbf{z}^*) = Q\,\boldsymbol{\mu}(\mathbf{z})$.

Lemma~\ref{lem:operator-invariance} shows that any such isometric pair induces identical observation-space causal matrices. Therefore, for any two parameterizations yielding the same adjacent-observation distribution, $\mathbf{C}^*_{\tilde{\mathbf{x}}}(\mathbf{z}^*) = \mathbf{C}_{\tilde{\mathbf{x}}}(\mathbf{z})$ at every common point.
\end{proof}

\bibliographystyle{IEEEtran}
\bibliography{refs}

\end{document}